\pdfoutput=1

\documentclass[11pt]{article}

\usepackage[preprint]{acl}

\usepackage{times}
\usepackage{latexsym}
\usepackage{amsmath}
\usepackage{amsfonts}
\usepackage{booktabs}
\usepackage{algorithm}
\usepackage{makecell}
\usepackage{algpseudocode}

\usepackage[T1]{fontenc}

\usepackage[utf8]{inputenc}

\usepackage{microtype}

\usepackage{inconsolata}

\usepackage{graphicx}

\newcommand\blfootnote[1]{%
  \begingroup
  \renewcommand\thefootnote{}\footnote{#1}%
  \addtocounter{footnote}{-1}%
  \endgroup
}

\title{SEEKR: Selective Attention-Guided Knowledge Retention for Continual Learning of Large Language Models}

\author{
 \textbf{Jinghan He\textsuperscript{1,2}},
 \textbf{Haiyun Guo\textsuperscript{1,2}$^\ast$},
 \textbf{Kuan Zhu\textsuperscript{1,2}$^\ast$},
 \textbf{Zihan Zhao\textsuperscript{5}},
\\
 \textbf{Ming Tang\textsuperscript{1}},
 \textbf{Jinqiao Wang\textsuperscript{1,2,3,4}$^\ast$}
\\
\\
 \textsuperscript{1}Foundation Model Research Center, Institute of Automation, Chinese Academy of Sciences
 \\
 \textsuperscript{2}School of Artificial Intelligence, University of Chinese Academy of Sciences
 \\
 \textsuperscript{3}Peng Cheng Laboratory, 
 \textsuperscript{4}Wuhan AI Research,
 \textsuperscript{5}Chongqing University
\\
 \small{
   \texttt{hejinghan2022@ia.ac.cn}, \texttt{\{kuan.zhu, haiyun.guo, jqwang\}@nlpr.ia.ac.cn}
 }
}

\begin{document}
\maketitle
\blfootnote{$^\ast$ Corresponding author.}
\begin{abstract}

Continual learning (CL) is crucial for language models to dynamically adapt to the evolving real-world demands. To mitigate the catastrophic forgetting problem in CL, data replay has been proven a simple and effective strategy, and the subsequent data-replay-based distillation can further enhance the performance. However, existing methods fail to fully exploit the knowledge embedded in models from previous tasks, resulting in the need for a relatively large number of replay samples to achieve good results. In this work, we first explore and emphasize the importance of attention weights in knowledge retention, and then propose a \textbf{SE}lective att\textbf{E}ntion-guided \textbf{K}nowledge \textbf{R}etention method (SEEKR) for data-efficient replay-based continual learning of large language models (LLMs). Specifically, SEEKR performs attention distillation on the selected attention heads for finer-grained knowledge retention, where the proposed forgettability-based and task-sensitivity-based measures are used to identify the most valuable attention heads. Experimental results on two continual learning benchmarks for LLMs demonstrate the superiority of SEEKR over the existing methods on both performance and efficiency. Explicitly, SEEKR achieves comparable or even better performance with only 1/10 of the replayed data used by other methods, and reduces the proportion of replayed data to 1\%. The code is available at \url{https://github.com/jinghan1he/SEEKR}.

\end{abstract}

\section{Introduction}

\begin{figure}[t]
\centering
  \includegraphics[width=\columnwidth]{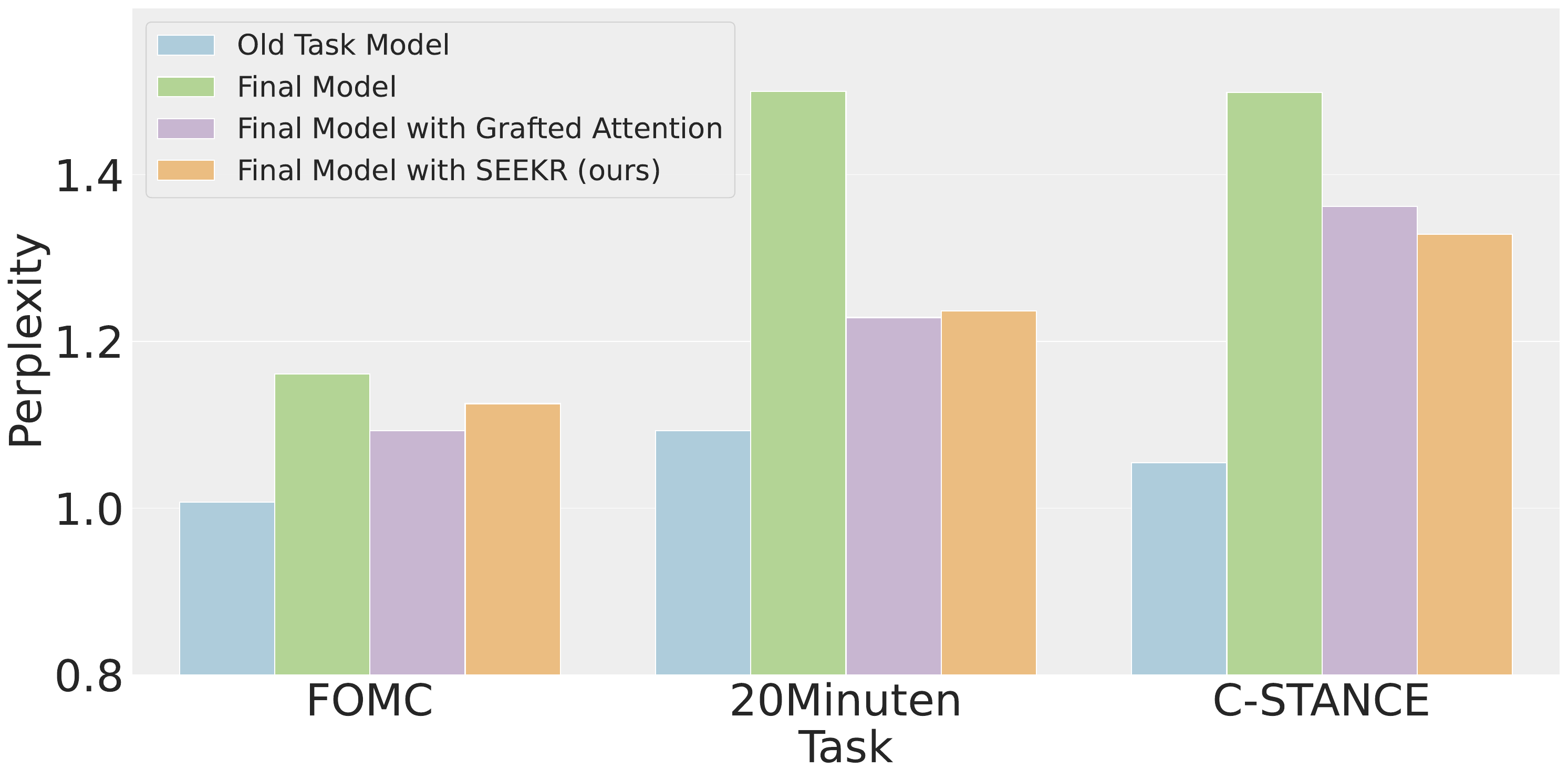}
  \caption{Demonstration of the critical role of attention weights in knowledge retention. We apply DER++ \cite{buzzega2020dark} for continual learning on the TRACE benchmark \cite{wang2023trace} to obtain multiple old task models and the final model. Grafting the attention weights of the old models onto the final model at inference can maintain better performance on the old tasks. Moreover, the final model obtained by our continual learning method, SEEKR, achieves similar results.}
  \label{fig:attn_graft}
\end{figure}

Enabling large language models~\cite{achiam2023gpt,touvron2023llama,zheng2024judging} with human-like continual learning ability is crucial for the long-term practical deployment. It allows for constant knowledge accumulation on new tasks without the need for costly retraining. However, sequentially finetuning the LLMs with new data can lead to catastrophic forgetting~\cite{mccloskey1989catastrophic}, impairing the general ability of the model and its performance on previous tasks.

Among the array of continual learning methods~\cite{ke2022continual}, data replay stands out as the most widely adopted strategy in practice due to its simplicity and efficacy~\cite{wang2024inscl}. Based on it, replay-based distillation methods, including DER++~\cite{buzzega2020dark} and subsequent techniques~\cite{qin2021lfpt5,kang2022class,gu2023preserving}, further boost the performance by utilizing memories from both data and model perspectives. Specifically, \citealp{buzzega2020dark,qin2021lfpt5,gu2023preserving} distill the output logits of old models for knowledge transfer, and \citealp{kang2022class} restrict the changes in important feature maps in the image encoders. However, these works have not fully exploited the potential of knowledge distillation in continual learning for LLMs. They focus on the outputs of network layers while neglecting the preservation of intricate internal functions. Consequently, a relatively large amount of replay data is required by these methods to achieve good results. 

Recently, many studies have investigated the attention weights of different heads to analyze the interpretability of the internal mechanisms in LLMs \cite{vig2019analyzing,wang2023label}. Inspired by this, we explore whether attention weights play a critical role in knowledge retention during continual learning in LLMs. As shown in Figure \ref{fig:attn_graft}, grafting the attention weights from the LLM of the old tasks to the final LLM after continual learning can maintain better performance on old tasks, which suggests that the attention weights could be crucial to alleviate the catastrophic forgetting problem and achieve more comprehensive knowledge retention\footnotemark. However, naively preserving the attention weights of all heads in the LLM by distillation introduces significant computational costs. Previous studies have observed a functional specialization phenomenon among attention heads in LLMs \cite{vig2019analyzing,jo2020roles,li2023interpreting},  which indicates the susceptibility of attention heads to forgetting and their importance to previous tasks vary. This property allows us to selectively focus on the valuable attention heads for efficient knowledge retention. 

\footnotetext[1]{Attention grafting can only be used during inference with both the source and target models, which is an infeasible solution for continual learning. We employ this technique solely for exploratory experiments.}

To this end, we propose a finer-grained model distillation method called \textbf{SE}lective att\textbf{E}ntion-guided \textbf{K}nowledge \textbf{R}etention (\textbf{SEEKR}) for continual learning of large language models, which employs attention distillation on the most valuable heads in LLMs to achieve efficient knowledge retention. Specifically, we develop knowledge-retention-oriented head importance measures, which consider both forgettability and task sensitivity, to identify the most valuable heads for distillation. The forgettability, measured by the cumulative changes in attention weights during continual learning, indicates the generality of knowledge and the necessity of distillation. An attention head with higher forgettability indicates a greater need for knowledge retention. The task sensitivity, calculated as the first-order derivative of the task loss, evaluates the importance of maintaining the attention weights of an attention head for a given task. An attention head with greater sensitivity should be prioritized to restrict variations in its attention weights. Using the above two importance scores, SEEKR designs a hierarchical budget allocation mechanism to adaptively select the most valuable attention heads for distillation in a controllable way, which can efficiently regulate the training cost. By using SEEKR, the performance of old tasks can be further maintained as shown in Figure \ref{fig:attn_graft}. 

Extensive experiments are conducted on the recently developed continual learning benchmark for LLMs \cite{wang2023trace} and the continual learning benchmark on traditional NLP tasks \cite{wang2022super}. The results consistently demonstrate the superiority of SEEKR in mitigating catastrophic forgetting and maintaining the general capabilities of LLMs. Moreover, as a replay-based method, SEEKR exhibits excellent data efficiency, achieving comparable or better performance with just 1/10 of the replayed data used by the existing methods, reducing the replayed data proportion to only 1\%. 

Our main contributions are summarized as follows:

\begin{itemize}
    \item We explore and emphasize the importance of attention weights for knowledge retention, and devise knowledge-retention-oriented measures to identify important attention heads for distillation. The proposed method, SEEKR, can efficiently preserve the finer-grained knowledge in the selected attention heads. 
    \item Extensive experiments validate the superiority of SEEKR, showcasing its data efficiency by using just 1\% of replay samples to achieve the comparable or better performance that other methods reach with 10\% of replay samples. 
\end{itemize}

\section{Preliminary}

\subsection{Continual Learning for LLMs}

Continual learning algorithms aim to accumulate knowledge across sequential tasks. Suppose there are $N$ tasks with the corresponding datasets $\{\mathcal D_1, \cdots, \mathcal D_N\}$. An LLM, parameterized by $\theta$, are instruction-tuned on each dataset $\mathcal D_i$ sequentially to optimize the following objective:
\begin{equation}
    L_{task}=\mathbb{E}_{(\boldsymbol{x},\boldsymbol{y})\in \mathcal{D}_i}\big[-\log p_{\theta}(\boldsymbol{y}| \boldsymbol x)\big]
\end{equation}

\noindent where $\boldsymbol{x}, \boldsymbol{y}$ are the instruction and true answer, respectively. Hereafter, we assume the current task is $i$ and omit the corresponding subscript. In this paper, we study a more common scenario in practice where a small amount of data from the old tasks $\{R_1, ..., R_N\}$ can be stored in the memory buffer to aid the continual learning process. During training on the current task, replay data are acquired from the memory buffer, and the model is optimized for their previous tasks:
\begin{equation}
    L_{replay}=\sum_{k=1}^{i-1} \mathbb{E}_{(\boldsymbol{x},\boldsymbol{y}) \in \mathcal{R}_k}\big[-\log p_{\theta}(\boldsymbol{y}| \boldsymbol x)\big]
\end{equation}

\subsection{Knowledge Distillation for CL}

Knowledge distillation~\cite{hinton2015distilling} is a technique to train a student model to replicate the teacher model's behavior for efficient knowledge transfer. To mitigate forgetting on previous tasks in CL, knowledge distillation is performed between each old model $p_{\theta_k}$ and the current model $p_{\theta}$ using replay samples from $R_k$ \cite{buzzega2020dark}:
\begin{equation}
  L_{ld} = \sum_{k=1}^{i-1} \mathbb{E}_{(\boldsymbol{x},\boldsymbol{y}) \in \mathcal{R}_k}\big[
D_{KL}(p_{\theta_k}(\boldsymbol{y}|\boldsymbol{x})\|p_\theta(\boldsymbol{y}|\boldsymbol{x})) \big]
\end{equation}

\noindent The predicted logits from the old model $p_{\theta_k}(\boldsymbol{y}|\boldsymbol{x})$ are saved in the memory buffer along with the replay samples and loaded during training as auxiliary supervision signals.

\section{Method}

In this section, we introduce SEEKR, an efficient replay-based distillation method that identifies valuable attention heads and performs attention distillation for finer-grained knowledge retention.

\subsection{Attention-guided Knowledge Retention}

To achieve more comprehensive knowledge retention by using less replay data, we perform an elaborate distillation on the key mechanism of LLMs, \textit{i.e.} the attention weights. Specifically, the outputted attention weights of the $h$-th head in the $l$-th layer are denoted as $A_{l,h}$ :
\begin{equation}
    A_{l,h}=\operatorname{softmax}(\frac{Q_{l,h}K_{l,h}^T}{\sqrt{d_k}} + M_{causal})
\end{equation}

\noindent where $Q$ and $K$ represent the query vectors and the key vectors in the self-attention operation, respectively. $M_{causal}$ is the casual attention mask in LLMs. We use $t$ to index the attention distribution of the $t$-th query in $A_{l,h}$ and denote it as $A_{l,h,t}$. The attention distributions of query $t$ from each old task model $A_{l,h,t}^k$ and the current model $A_{l,h,t}$ are aligned through the KL divergence loss:
\begin{equation}
L_{ad}(A, A^k) = \sum_{(l,h)\in U}\sum_{t=1}^{|\boldsymbol{x}\oplus \boldsymbol{y}|}{D_{KL}(A_{l,h,t}^k \| A_{l,h,t})}
\label{eq:ad}
\end{equation}
\noindent where $U$ stands for the set of all attention heads in all layers. $\boldsymbol{x}\oplus \boldsymbol{y}$ is the concatenated sequence of $\boldsymbol{x}$ and $\boldsymbol{y}$, and $|\boldsymbol{x}\oplus \boldsymbol{y}|$ means the length of the whole sequence. In SEEKR, the knowledge distillation is performed at the head level, which can offer more direct and refined regulation on the intricate internal functions of LLMs, achieving a more comprehensive and efficient utilization of the limited replay data. 

\subsection{Important Head Identification}

In practice, distilling all the attention heads in an LLM is costly and unnecessary, as different heads exhibit varying levels of task sensitivity and forgettability. Therefore, we propose a two-dimensional measure to identify the most valuable attention heads for knowledge retention.

\subsubsection{Task Sensitivity Measure}

For a model adapted to task $k$, we assess to which extent changes in the attention weights of each head affect the task performance. Following common practice, we resort to Taylor expansion to formalize this influence~\cite{kang2022class}:
\begin{equation}
\begin{aligned}
    \Delta L(\boldsymbol{x},\boldsymbol{y}) &\approx \left\langle\frac{\partial L(\boldsymbol{x},\boldsymbol{y})}{\partial A_{l,h}}, \Delta A_{l,h}\right\rangle_F \\
    & \leq ||\frac{\partial L(\boldsymbol{x},\boldsymbol{y})}{\partial A_{l,h}}||_F \cdot ||\Delta A_{l,h}||_F
\end{aligned}
\end{equation}

\noindent where $\langle\cdot,\cdot\rangle_F$ and $\|\cdot\|_F$ denote the Frobenius inner product and  Frobenius norm, respectively. This inequality demonstrates the upper bound on the increase in task loss due to changes in the attention weights, \textit{i.e.} $\Delta A_{l,h}$. A larger coefficient indicates a higher upper bound for the same changes in $A_{l,h}$. This implies that changes in these attention weights are more likely to increase task loss or degrade task performance, making it crucial to keep them unchanged. Therefore, we take the coefficient to estimate the sensitivity of the task $k$ to $A_{l,h}^k$, which is formulated as:
\begin{equation}
    S^k_{l,h} = \mathbb{E}_{(\boldsymbol{x},\boldsymbol{y}) \in R_k}|| \frac{\partial L(\boldsymbol{x},\boldsymbol{y})}{\partial A^k_{l,h}}||_F
    \label{eq:sensitivity}
\end{equation}

The importance scores are then normalized within each layer to obtain $\widetilde S^k_{l,h}$, thereby mitigating the impact of varying gradient magnitudes across different layers. During training on the new task, the importance of all previous tasks should be considered. Therefore, the task sensitivity measure for each attention head is defined as:
\begin{equation}
    S_{l,h}=\sum_{k=1}^{i-1}\widetilde S_{l,h}^k
\label{eq:tasksens}
\end{equation}

\subsubsection{Forgettability Measure}
\label{sec:forgettability}
The second measure assesses the necessity for performing attention distillation on each attention head. We hypothesize that there exist some attention heads whose attention weights remain relatively stable during continual training on new tasks, suggesting that they are less sensitive to task-specific details and focus more on general or shared knowledge. This hypothesis aligns with prior research \cite{zhao2023does}, which revealed that only a few modules change drastically during continual learning, while others stay relatively stable and may be shared across tasks as common knowledge. Based on this, we propose that stable attention heads may encode general knowledge that is less prone to forgetting, and thus distillation of such heads should be minimized. To this end, we leverage the variability of the attention weights during continual learning to measure the forgettability of the attention head: 
\begin{equation}
    F_{l,h} = \sum_{k=1}^{i-1}\mathbb{E}_{(\boldsymbol{x},\boldsymbol{y})\in R_k} ||A_{l,h}^{k}-A^{k-1}_{l,h}||_F
    \label{eq:forgettability}
\end{equation}

Higher forgettability scores indicate a greater necessity for distilling these attention heads.

\subsubsection{Overall Importance Measure}

To identify valuable heads for attention-guided knowledge retention, we fuse the two complementary measures through multiplication, ultimately forming a holistic metric:
\begin{equation}
    I_{l,h} = S_{l,h} \cdot F_{l,h}
    \label{eq:importance}
\end{equation}
After each task, $S_{l,h}$ and $F_{l,h}$ of each attention head are updated according to Equation \ref{eq:tasksens} and \ref{eq:forgettability}, and the overall importance $I_{l,h}$ is re-calculated accordingly.

\begin{algorithm}[t]
\caption{SEEKR}
 \textbf{Input} Initial model $\theta_0$, Datasets $\{\mathcal{D}_i\}_{i=1}^{N}$, Hyperparameters $\lambda_1,\lambda_2,B_L,B_H,B_T$
\begin{algorithmic}[1]
    \State Initialize $L, H\leftarrow U;\  S_{l,h}, F_{l,h}, I_{l,h} \leftarrow 0$;
    \For{task $i$ $\leftarrow$ 1 to $N$}
     \For{epoch $e \leftarrow$ 1 to $epochs$}
     \For{batch in $(\bigcup_{k=1}^{i-1} R_k) \bigcup \mathcal{D}_i$}
     \State Minimize $L$ in Eq. \ref{eq:overall};
     \EndFor
     \EndFor
     \State $R_i \leftarrow \operatorname{Random}(\mathcal{D}_i)$;
     \State Update $S_{l,h},F_{l,h},I_{l,h}$ using Eq. \ref{eq:tasksens}-\ref{eq:importance};
     \State Update $L, H$ using Eq. \ref{eq:budget};
     \State Randomly select $T$;
   \EndFor
\end{algorithmic}
\label{alg:seekr}
\end{algorithm}

\subsection{Hierarchical Budget Allocation}

Based on the above head importance measure, we propose a hierarchical budget allocation strategy to manage the training cost. We define the group of selected layers and heads as $L$ and $H$, with budgets $B_L$ and $B_H$. Our strategy involves two steps: (1) Select the top-$B_L$ layers that maximize the layer-wise importance scores $\sum_h I_{l,h}$. (2) Among all the attention heads in all these layers, activate the top-$B_H$ heads for attention distillation. Based on the above process, the set $H$ of the selected heads can be expressed as:
\begin{equation}
\begin{aligned}
    H&=\arg\mathop{\mathrm{topk}}_{(l,h)} \{I_{l,h}\ |\  l \in L\} \\
    L&=\arg\mathop{\mathrm{topk}}_{l}\sum_h I_{l,h}
\end{aligned}
\label{eq:budget}
\end{equation}
where $\arg\mathop{\mathrm{topk}}_{z}$ denotes the set of $z$ that achieves the $k$ largest values. $k$ is $B_H$ for $H$ and $B_L$ for $L$. Additionally, to reduce the $O(n^2)$ cost of distilling the entire attention map, we introduce a query budget $B_T$ and randomly select the queries $T$ for distillation. After determining $H$ and $T$, we can rewrite Equation \ref{eq:ad} as follows:
\begin{equation}
\begin{aligned}
L_{ad}(A, A^k) &= \sum_{(l,h)\in H}\sum_{t\in T}{D_{KL}(A_{l,h,t}^k \| A_{l,h,t})} \\
L_{seekr} &= \sum_{k=1}^{i-1} \mathbb{E}_{(\boldsymbol{x},\boldsymbol{y}) \in \mathcal{R}_k}\big[ L_{ad}(A,A^k)\big]
\end{aligned}
\end{equation}

Overall, SEEKR sets three types of budgets to allow flexible control over training costs. First, the layer budget adjusts the number of layers for attention-accelerating algorithms or our distillation strategy. Second, the head budget filters out less essential heads and reduces training costs. Lastly, the query budget specifically targets at reducing the costs associated with distilling long texts.

\begin{table*}[t]
  \centering
  \begin{tabular}{l|cc|cc}
    \toprule
    & \multicolumn{2}{c}{\textbf{LLaMA-2-7B-Chat}}& \multicolumn{2}{|c}{\textbf{Vicuna-7B-v1.5}}\\
    & \textbf{Order1}&                            \textbf{Order2}& \textbf{Order1}&\textbf{Order2}\\
    \midrule
    SeqFT& 47.63 (-11.45)&                            45.12 (-12.27)& 41.91 (-15.29)&45.70 (-12.01)\\
 EWC& 48.20 (-9.48)& 44.54 (-12.00)& 41.88 (-15.57)&49.32 (-8.62)\\
 LwF& 41.86 (-6.50)& 40.25 (-5.96)& 41.19 (-5.54)&42.99 (-4.72)\\
 LFPT5& 38.67 (-11.43)&  42.26 (-7.43)& 41.79 (-8.10)&39.22 (-10.70)\\
 L2P& 35.23 (-15.96)& 34.63 (-16.86)& 32.26 (-16.58)&35.14 (-15.88)\\
    PP& 29.41 (-5.79)&                            21.58 (-8.83)& 26.64 (-6.10)&24.88 (-11.54)\\
 O-LoRA& 44.64 (-4.20)& 42.83 (-9.11)& 43.42 (-6.27)&43.87 (-6.37)\\
 \midrule
 Replay (1\%)& 48.47 (-9.69)& 47.04 (-10.24)& 48.43 (-9.23)&49.46 (-9.43)\\
 DER++ (1\%)& 49.22 (-8.32)& 46.59 (-10.91)& 49.01 (-9.04)&51.09 (-7.85)\\
 \textbf{SEEKR} (1\%)& \textbf{54.99} (\textbf{-2.61})& \textbf{54.69} (\textbf{-2.53})& \textbf{55.78} (\textbf{-2.64})&\textbf{54.91} (\textbf{-3.40})\\
 \midrule
 Replay (10\%)& 55.67 (-3.96)& 53.39 (-4.15)& 55.62 (-2.15)&54.57 (-3.41)\\
 DER++ (10\%)& 55.01 (-3.50)& 54.05 (-2.94)& 56.06 (-1.17)&55.14 (-3.77)\\
 \textbf{SEEKR} (10\%)& \textbf{58.27} (\textbf{0.11})& \textbf{57.27} (\textbf{-0.47})& \textbf{57.54} (\textbf{0.47})&\textbf{56.86} (\textbf{-1.01})\\
 \midrule
    MTL&                         \multicolumn{2}{c|}{59.38}& \multicolumn{2}{c}{58.18}\\
    \bottomrule
  \end{tabular}
  \caption{Comparison with the state-of-the-art methods on TRACE benchmark. The results are obtained by using two popular LLMs with two transfer orders, and are presented in the format of OP (BWT).
  }
\label{tab:mainresults}
\end{table*}

\subsection{Overall Objective}

Combining the above objectives, the overall loss for the new and replay data is formalized as:
\begin{equation}
\label{eq:overall}
    L =L_{task} + \lambda_1 L_{replay} + (1-\lambda_1) L_{ld} + \lambda_2 L_{seekr}
\end{equation}

\noindent where $\lambda_1$ is a coefficient to balance the text generation loss supervised by true labels and teacher models, and $\lambda_2$ is a weighting factor to adjust the magnitude of attention distillation loss. The overall process of SEEKR is shown in Algorithm \ref{alg:seekr}.

\section{Experiments}

\begin{table*}[t]
  \centering
\begin{tabular}{l|cccccc|c}
\toprule
    & \textbf{MMLU}&                            \textbf{GSM}& \textbf{BBH}&\textbf{TydiQA}&\textbf{BoolQ} &\textbf{PIQA} & \textbf{GA (DeltaGA)}\\
    \midrule
    LLaMA-2-7B-Chat& 46.89 &                            27.14& 39.73 &16.76 & 79.79& 76.33&47.77 
\\
 SeqFT& 45.16 & 14.03 & 32.50 &14.84 & 79.00 & 75.49 &43.50 (-4.27)\\
 Replay (1\%)& 45.49 & 12.70 & 33.46 & 14.65 & 78.69 & 75.65 &43.44 (-4.33)\\
 \textbf{SEEKR} (1\%)& 46.32 & 20.85 & 38.52 &18.22 & 80.64 & 75.79 &\textbf{46.72 (-1.05)}\\
 \midrule
 Vicuna-7B-v1.5& 49.39 &                            23.43& 41.12 &15.01 & 81.41& 76.77&47.86 
\\
 SeqFT& 46.26 & 11.68 & 33.09 &13.44 & 79.97 & 76.72 &43.52 (-4.34)\\
 Replay (1\%)& 47.14 & 15.77 & 33.51 & 14.14 & 80.57 & 76.39 &44.59 (-3.27)\\
 \textbf{SEEKR} (1\%)& 48.83 & 17.55 & 38.17 &16.32 & 81.96 & 77.23 &\textbf{46.68 (-1.18)}\\
 \bottomrule
  \end{tabular}
  \caption{
   Changes in general language understanding and reasoning abilities after continual learning with different methods. The reported results of all continual learning models are averaged over two task orders.
  }
\label{tab:general}
\end{table*}

\subsection{Experimental Setup}

\subsubsection{Datasets}

~~~\textbf{CL Benchmark for LLMs.} 
We evaluate our method on TRACE  \cite{wang2023trace}, a continual learning benchmark for LLMs that includes eight datasets covering domain-specific knowledge, multilingual capabilities, code generation, and mathematical reasoning. We use the reasoning-augmented version of datasets and conduct experiments under two task orders following the original paper. After continual learning, we assess the performance of the continually learned tasks and the changes in the general ability of LLMs. 

\textbf{CL on Traditional NLP Tasks.} SuperNI \cite{wang2022super} contains a variety of traditional NLP tasks and can serve as a practical benchmark for continual learning of large language models. Similar to \citealp{zhao2024dapt}, we select three datasets for each of the four types of tasks, \textit{i.e.} information extraction, question answering, summarization, and sentiment analysis, to examine the effectiveness of continual learning methods. For each dataset, 1000 samples and 100 samples are randomly sampled for training and testing, respectively.

\subsubsection{Metrics}

Let $a_{i,j}$ denote the testing performance on the $i$-th task after training on the $j$-th task. We report the overall performance (OP)~\cite{chaudhry2018riemannian} and the backward transfer (BWT)~\cite{lopez2017gradient} after training on the last task:
\begin{equation}
    OP = \frac{1}{T} \sum_{i=1}^T a_{i,T}
\end{equation}
\begin{equation}
    BWT = \frac{1}{T-1} \sum_{i=1}^{T-1} (a_{i,T}-a_{i,i})
\end{equation}
 \noindent Moreover, we also report the general ability (GA) and the delta general ability (DeltaGA)~\cite{wang2023trace} after continual learning. GA is the average performance across evaluation datasets in Table \ref{tab:general} and DeltaGA shows the change in GA compared to the initial model.

\subsection{Baselines}

We compare SEEKR with nine baseline methods: (1) \textbf{SeqFT} sequentially finetunes the model without continual learning strategies. (2) \textbf{EWC}~\cite{kirkpatrick2017overcoming} regularizes parameter variations based on parameter importance scores. (3) \textbf{LwF}~\cite{li2017learning} distills the model of the last task using the current task data. (4) \textbf{Replay} finetunes the model with the current task data and a small number of replay samples. (5) \textbf{DER++}~\cite{buzzega2020dark} saves the logits of the replay samples from the old models for distillation, and combines distillation and replay to reduce forgetting. (6) \textbf{LFPT5}~\cite{qin2021lfpt5} learns a soft prompt to generate pseudo samples of previous tasks for replaying. (7) \textbf{O-LoRA}~\cite{wang2023orthogonal} imposes orthogonal constraints on the LoRA matrices for all tasks. (8) \textbf{L2P}~\cite{wang2022learning} instantiates a prompt pool for adaptive prompt selection and prompt tuning for individual samples. (9) \textbf{PP}~\cite{razdaibiedina2023progressive} tunes a set of prompts for each task and concatenates them together. In addition, the results of the multi-task trained models are reported as \textbf{MTL} and serve as the upper-bound reference. 

\begin{figure*}[t]
\centering
  \includegraphics[width=0.9\linewidth]{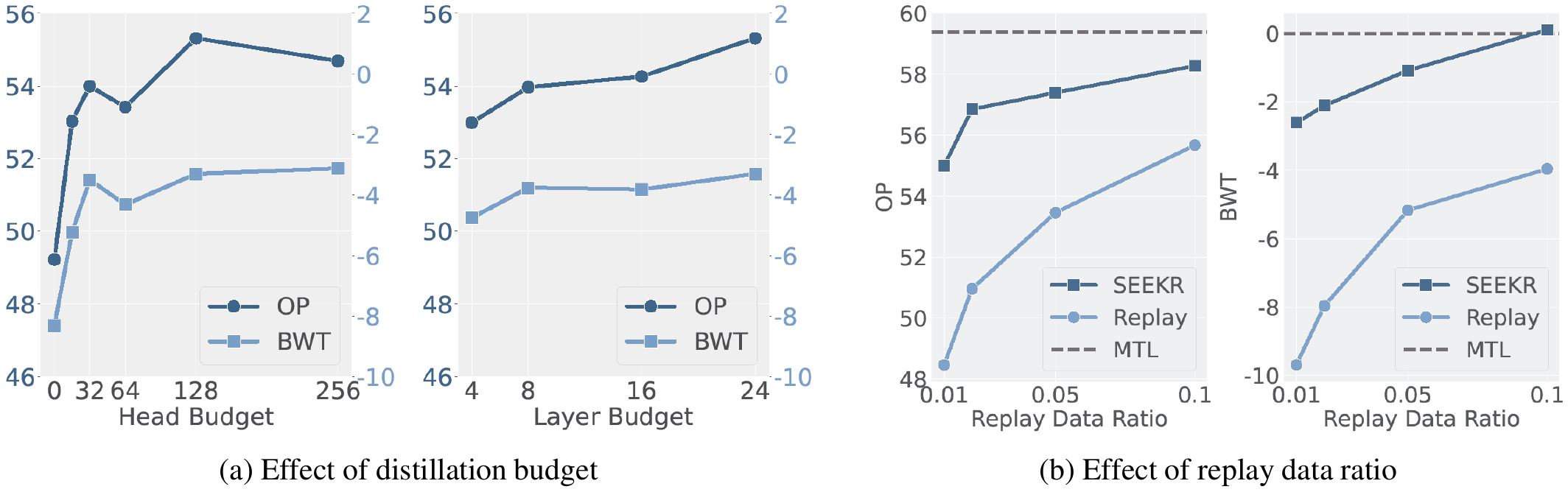}
  \caption{Results of SEEKR across different distillation budgets and different replay data ratios.  }
  \label{fig:budget}
\end{figure*}

\subsection{Implementation Details}

SEEKR is a versatile continual learning method compatible with any transformer-based model. Following ~\citealp{wang2023trace}, we conduct our main experiments on two popular LLMs, \textit{i.e.} LLaMA-2-7B-chat~\cite{touvron2023llama} and Vicuna-7B-v1.5~\cite{zheng2024judging}. We also scale to a larger model Vicuna-13B-v1.5 to validate the effectiveness of SEEKR. All models are trained on 8 NVIDIA Tesla A800 using the DeepSpeed library. The training batch size is 128. For methods not involving parameter-efficient tuning modules, the learning rate is 1e-5. For replay-based methods, the default replay ratio is 1\%. For SEEKR, $\lambda_1$ in Equation~\ref{eq:overall} is set to 0.5. $\lambda_2$ is 1e3 for a replay ratio of 1\% and 1e2 for 10\%. The head budget $B_H$ is 128, and the layer budget $B_L$ is 24 by default and 8 for 13B models or a replay ratio of 10\%. The query budget $B_T$ is 100. All experimental results were averaged over 3 runs. More implementation details can be found in Appendix \ref{sec:impl}.

\begin{table}[t]
  \centering
  \small
  \begin{tabular}{l|cc}
    \toprule
    & \textbf{Order3}&\textbf{Order4}\\
    \midrule
    SeqFT& 42.62 (-18.12) &50.52 (-9.88)\\
    LwF& 43.29 (-15.47) &47.35 (-12.57)\\
 LFPT5&42.05 (-16.26) &46.09 (-14.16)\\
 L2P&32.71 (-22.34) &31.00 (-23.82)\\
 PP& 17.96 (-21.27)&12.19 (-29.08)\\
 O-LoRA&30.07 (-24.47) &26.70 (-33.82)\\
 Replay (1\%)&55.00 (-4.27) &54.78 (-5.31)\\
 DER++ (1\%)&55.89 (-4.51) &53.48 (-5.01)\\
 \textbf{SEEKR} (1\%)&\textbf{57.04} (\textbf{-3.15})&\textbf{58.26} (\textbf{-2.52})\\
 \midrule
    MTL& \multicolumn{2}{c}{61.27}\\
    \bottomrule
  \end{tabular}
  \caption{Comparison with the state-of-the-art methods on SuperNI benchmark. The experiments are conducted on LLaMA-2-7B.}
  \label{tab:superni}
\end{table}

\subsection{Main Results}

Table \ref{tab:mainresults} compares the overall continual learning performance of SEEKR with other baselines on TRACE benchmark. Following \citealp{wang2023trace}, we also report the changes in the general ability of LLMs after continual learning in Table \ref{tab:general}. Similar experiments on the SuperNI benchmark are displayed in Table \ref{tab:superni}. 

\textbf{SEEKR effectively mitigates catastrophic forgetting of continually learned tasks.}  Compared to traditional and state-of-the-art continual learning approaches, SEEKR consistently achieves the highest OP and the lowest magnitude of BWT in all settings. Note that the BWT metric specifically captures the resistance of methods to catastrophic forgetting, thus the results demonstrate SEEKR's superiority in maintaining performance on newly learned tasks. Additionally, on the SuperNI benchmark, we achieve the best performance using only a small proportion of replay samples, likely because the benchmark consists of traditional NLP tasks, which are less challenging.

\textbf{SEEKR fully exploits the small amount of replay data and exhibits excellent data efficiency.} Among all replay-based methods, SEEKR stands out with a distinct advantage. On the TRACE benchmark, both Replay and DER++ show limited benefits with a lower ratio of replay data. In contrast, SEEKR demonstrates remarkable performance with just 1\% of the samples replayed, achieving comparable or even better results than other methods that replay 10\% of the samples. This underscores the ability of SEEKR to maximize the use of a small number of old samples and the inherent knowledge in the old models. 

\textbf{SEEKR is effective in maintaining the general ability of the original LLM.} Table \ref{tab:general} exhibits the changes in LLMs' general ability after continual learning. LLMs that are continually trained on new tasks show a decline in general task performance, demonstrating the catastrophic forgetting of their original capabilities.  Results validated that SEEKR, which elaborately distills multiple finetuned LLMs with a variety of data, helps to maintain the general capabilities of the model. This could benefit from the fact that our approach preserves the knowledge of the intricate internal functions in LLMs at the attention head level. 

\begin{table}[t]
  \centering
  \small
  \begin{tabular}{lcc}
    \toprule
    &  \textbf{Order1}&\textbf{Order2}\\
    \midrule
    random&  53.25 (-4.63)&52.62 (-5.11)\\
    task-sensitivity-only&  53.91 (-4.29)&53.56 (-2.84)\\
    forgettability-only&  54.06 (-3.31)&53.63 (-3.11)\\
    both&  \textbf{54.99} (\textbf{-2.61})&\textbf{54.69} (\textbf{-2.53})\\
    \bottomrule
  \end{tabular}
  \caption{Ablation study on the head importance measure. The experiments are conducted on LLaMA-2-7B.}
  \label{tab:ablation}
\end{table}

\begin{figure}[t]
\centering
  \includegraphics[width=0.9\columnwidth]{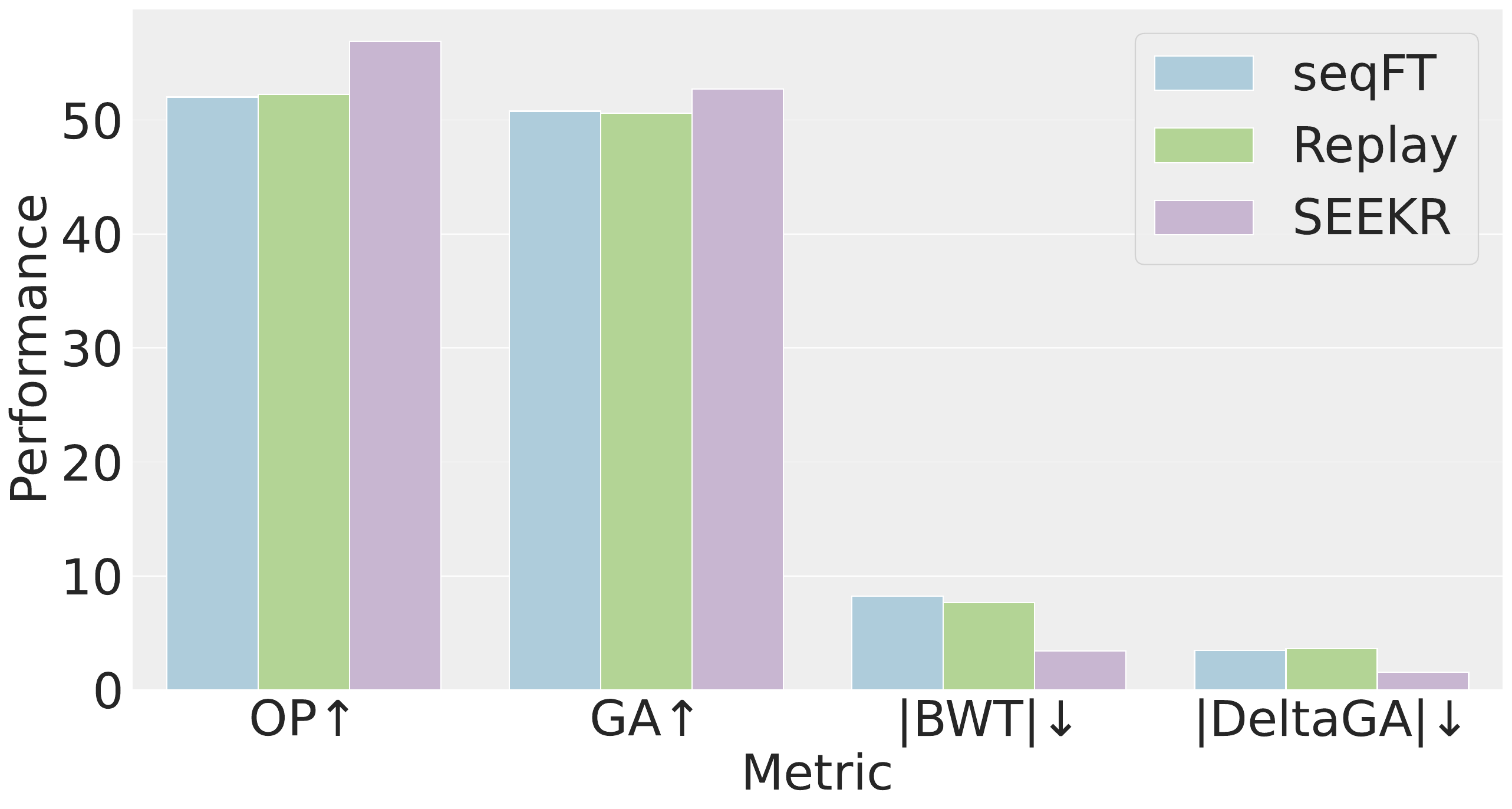}
  \caption{The continual learning performance and the changes of general ability with Vicuna-13B-v1.5.  }
  \label{fig:13b}
\end{figure}

\subsection{Ablation Studies}

~~~\textbf{Effect of distillation budget.} 
Figure \ref{fig:budget} (a) exhibits the performance of our method under different budgets. With a fixed layer budget of 24, a larger head budget can lead to better results, but this improvement tends to plateau at a budget of 128. Similarly, the performance improves with an increasing layer budget and reaches its optimum at 24. These results further emphasize the significance of distilling the right attention heads. Distilling less essential attention heads may lead to ineffective work.

\textbf{Effect of more replay samples.}
To further explore the potential of SEEKR, we experiment with an increased ratio of replay samples. Meanwhile, we compare SEEKR with Replay to demonstrate its data efficiency. As shown in Figure \ref{fig:budget} (b), SEEKR steadily improves performance as the number of replay samples grows. At a replay ratio of 10\%, the BWT score exceeds 0, indicating no forgetting or even a positive transfer has been achieved, and the overall performance approximates the upper bound of multi-task training. Moreover, compared with Replay, SEEKR is very data efficient by utilizing only 1\% of the old data to achieve the performance of replaying ten times that amount.

\textbf{Effectiveness of our head importance measure.}
We present the results of the ablation study on the proposed head importance measure in Table \ref{tab:ablation} . The results show that the random selection of distilled attention heads noticeably resulted in a higher forgetting indicator, while using either sensitivity-based or variation-based measures helps identify important heads for knowledge retention. Finally, combining both of the above measures produces the best results.

\begin{figure}[t]
\centering
  \includegraphics[width=0.9\columnwidth]{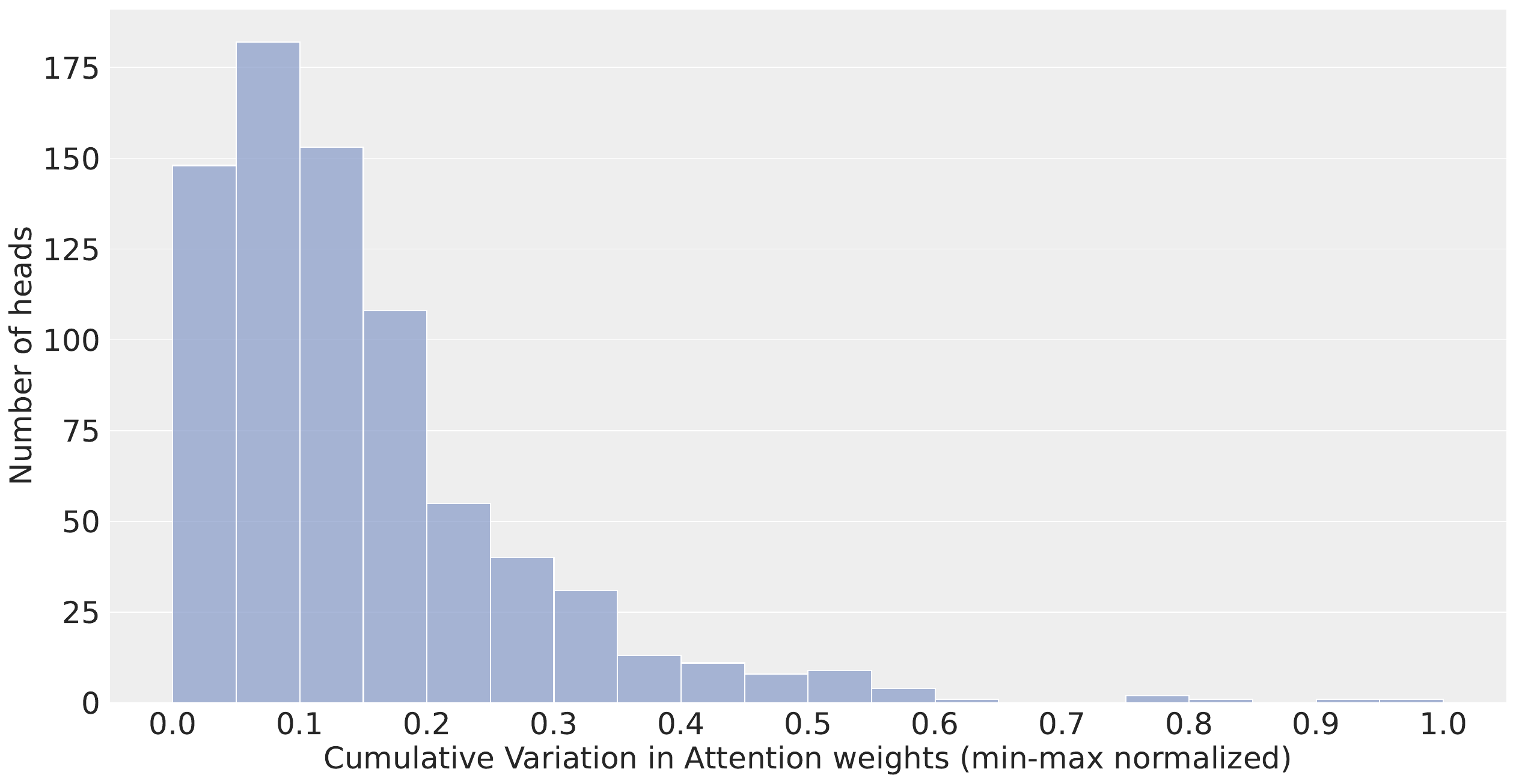}
  \caption{Histogram of the cumulative variation in the attention weights of the attention heads in the model during sequential finetuning.}
  \label{fig:attn_diffs}
\end{figure}

\begin{figure}[t]
\centering
  \includegraphics[width=0.9\columnwidth]{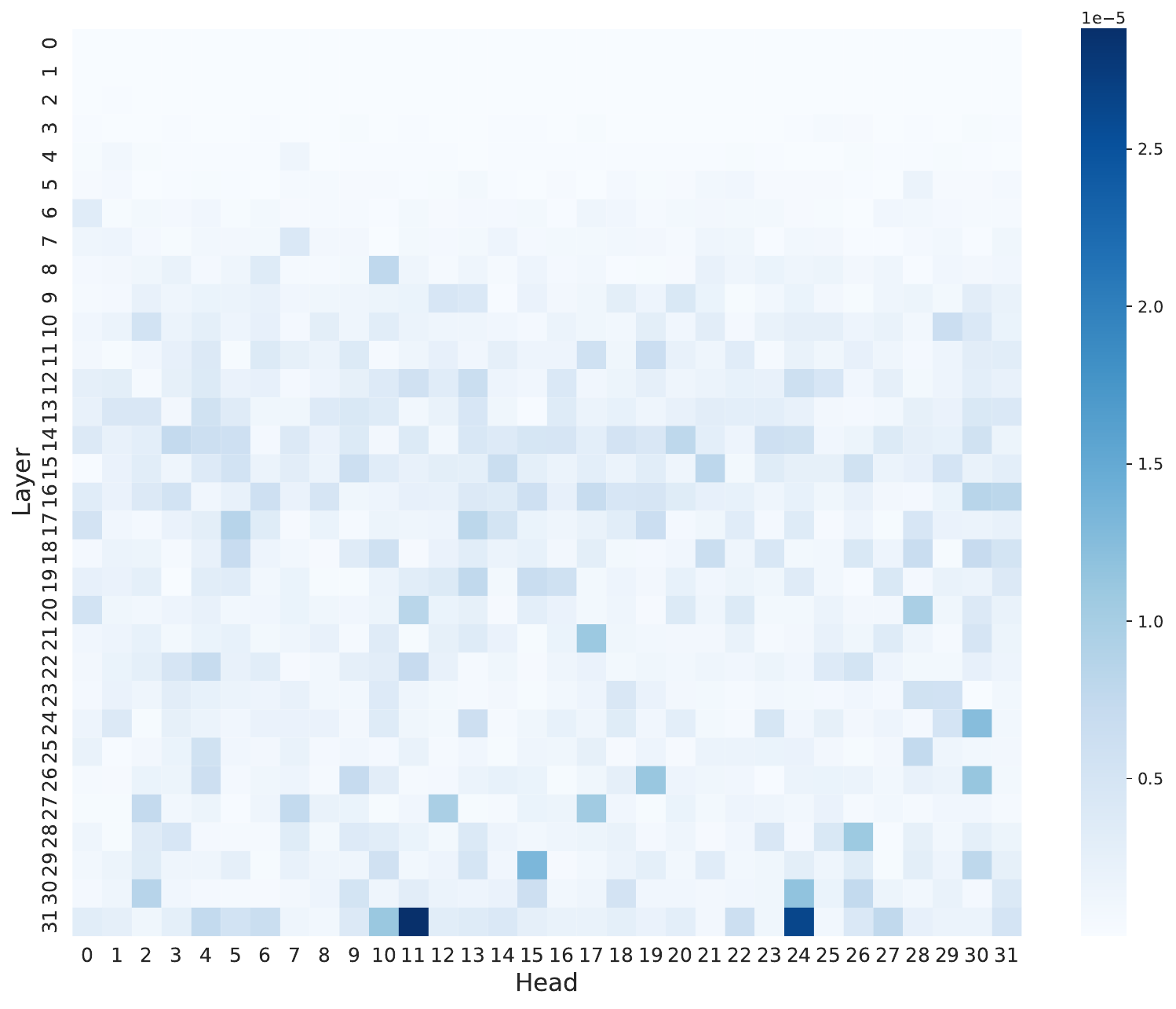}
  \caption{Visualization of the importance scores of all heads in the model.}
  \label{fig:attn_ipt}
\end{figure}

\subsection{Discussions}

~~~\textbf{Scale to larger models.} To validate the generalizability of SEEKR across different model scales, we conducted additional experiments on a larger model, Vicuna-13B-v1.5. Figure \ref{fig:13b} shows that our approach still effectively preserves both the performance of newly learned tasks and the general capabilities of the original model. 

\textbf{Variation in attention weights.} To further confirm our hypothesis in Section \ref{sec:forgettability}, we examine the cumulative changes in attention weights of each attention head during sequential finetuning. The results in Figure \ref{fig:attn_diffs} reveal that most attention heads remain stable throughout the process, while a small proportion undergo significant changes. This observation is similar to prior findings \cite{zhao2023does} and supports our hypothesis that these stable attention heads do exist, making it reasonable to identify them and avoid unnecessary attention distillation.

\textbf{Analysis of selected important heads.}
Figure \ref{fig:attn_ipt} illustrates that important attention heads are mainly distributed in the middle and deep layers of the model, while almost none are observed in the shallow layers. This aligns with the idea that the shallow layers encode more generalized knowledge and are less susceptible to forgetting. A closer look at Figure \ref{fig:attn_ipt} further reveals that the importance scores for the deeper layers are concentrated in a few heads, while those for the middle layers are more evenly spread over a larger number of heads. This may be because the heads in the deeper layers are more thoroughly function-specialized.

\section{Related Works}

\subsection{Continual Learning for LLMs}

Existing continual learning methods are typically classified into three broad categories: regularization-based methods, replay-based methods, and architectural-based methods. (1) \textbf{Regularization-based methods} restrict model variations to alleviate forgetting. Some works penalize changes to important parameters for previously learned tasks \cite{kirkpatrick2017overcoming,wang2023orthogonal,he2023continual}, while others resort to knowledge distillation to maintain the old models' predictions \cite{li2017learning,buzzega2020dark,kang2022class}. (2) \textbf{Replay-based methods} replay data from the old tasks during training on the new task. Experience replay methods \cite{rebuffi2017icarl,wang2024inscl} design data selection strategies of previous samples, and generative replay \cite{shin2017continual,qin2021lfpt5} uses generative models to produce synthetic data from previous tasks. Other methods \cite{yang2023continual} retain old tasks by storing statistical information of the old tasks instead of the original data. (3) \textbf{Architecture-based methods} alter the model structure to accommodate different tasks. Recently, this type of methods on LLMs \cite{wang2022learning,razdaibiedina2023progressive} often add parameter-efficient tuning modules for new tasks. 

SEEKR falls into the category of replay-based distillation methods and focuses on the preservation of important attention mechanisms in LLMs. Unlike existing output or parameter importance measures \cite{kirkpatrick2017overcoming,kang2022class}, which focus solely on task loss sensitivity, our head importance measure includes a forgettability aspect. This reflects the susceptibility to forgetting and the generality of knowledge in different heads, thereby determining the necessity for distillation. 

\subsection{Knowledge Distillation}

Knowledge distillation aims to leverage the teacher model’s performance and generalize it to the student model \cite{hinton2015distilling,park2019relational,guo2023semantic}. For language models, \citealp{sanh2019distilbert} uses the teacher model's generation distribution for each token as a supervision signal for the student model, and some other works \cite{wang2020minilm,wang2020minilmv2} distill the attention scores of one layer to transfer the knowledge of larger LMs into smaller models. Unlike their objectives of transferring knowledge between models of different sizes, we use attention distillation for knowledge retention. Both our teacher and student models share a similar architecture and are derived from the same pre-trained LLM, which enables head-by-head and layer-by-layer distillation.

\section{Conclusion}

In this paper, we propose SEEKR, an efficient replay-based distillation method for continual learning in LLMs. SEEKR resorts to attention distillation of important heads for finer-grained knowledge retention, which identifies valuable heads through the proposed knowledge-retention-oriented importance measures. Combined with a hierarchical budget allocation mechanism, SEEKR can ensure its utility across various resource levels. Extensive experiments consistently validated the effectiveness of our method in preserving the performance of newly learned tasks and the original ability of the initial LLMs. 
\section*{Limitations}

Despite the potential benefits of SEEKR, several limitations need to be considered. First, SEEKR is inherently a replay-based approach, which may not be applicable in scenarios where historical data involves privacy concerns. A potential solution is to use SEEKR with pseudo-samples generated by the trained LLM, but this approach requires further exploration. Second, due to computational resource limitations, we did not experiment with larger-scale LLMs like LLaMA-2-70B. Additionally, the application of SEEKR to continual learning with multimodal large language models remains to be explored in the future.  
\section*{Acknowledgements}

This work was supported by National Key R\&D Program of China under Grant No.2021ZD0110400, also partly supported by Beijing Natural Science Foundation under Grant 4244099, National Natural Science Foundation of China under Grant No.62276260, Postdoctoral Fellowship Program of CPSF under Grant GZC20232996, China Postdoctoral Science Foundation under Gant 2024M753498.

\bibliography{custom}

\appendix

\clearpage

\section{Datasets}

For the TRACE benchmark \cite{wang2023trace}, we conduct experiments on the reasoning-augmented datasets as such high-quality training data is more suitable for the LLM learning paradigm. The task order is consistent with the two orders provided by the benchmark, which are also displayed in Table \ref{tab:task_order}. For evaluation on the changes in the general ability, we test the LLMs on the datasets \cite{hendrycks2020measuring,ghazal2013bigbench,clark2020tydi,bisk2020piqa,clark2019boolq,cobbe2021training} included in this benchmark.

For the SuperNI benchmark \cite{wang2022super}, we choose four types of tasks and three dataset each for continual learning, containing a total of 12 traditional NLP tasks similar to \citealp{zhao2024dapt}. The two task orders can be found in Table ~\ref{tab:task_order}. 

\section{Implementation Details}
\label{sec:impl}

For methods not involving parameter-efficient tuning (PET) modules, we finetuning the LLMs on the task sequence in order1 for 5, 5, 5, 5, 5, 5, 10, 5 epochs, order2 for 10, 10, 10, 5, 5, 5, 5, 5 epochs, and order3 and order4 for 10 epochs each. For the compared baseline methods involving PET modules, the training epochs vary from 5 to 15 epochs for better performance. The hyperparameters of the compared baseline methods were kept the same as in the original repositories. If they did not perform well, we conducted additional searches for the optimal learning rate. 

For all the replay-based methods, we randomly selected the indicated proportion of replay samples from the full training set and kept the replay samples utilized by each method consistent for fairness. For the replay-based distillation methods, the distillation signals, \textit{i.e.} output logits and attention weights, of each old teacher model are saved in the memory buffer along with the original replay samples and loaded from the buffer during training on the new task. When replaying the old data, samples from the memory buffer and the current task are sampled in an evenly interleaved manner according to the ratio of their volumes. 

\begin{table*}
    \centering
\small    \begin{tabular}{c|c|c}
    \toprule
         \textbf{Order}&  \textbf{Benchmark}& \textbf{Task Sequence}\\
         \midrule
         1&  TRACE benchmark& \makecell{C-STANCE $\rightarrow$ FOMC $\rightarrow$ MeetingBank $\rightarrow$ Py150 $\rightarrow$ \\ScienceQA $\rightarrow$ NumGLUE-cm $\rightarrow$ NumGLUE-ds $\rightarrow$ 20Minuten}\\
 \midrule
 2& TRACE benchmark& \makecell{NumGLUE-cm $\rightarrow$ NumGLUE-ds $\rightarrow$ FOMC $\rightarrow$ 20Minuten $\rightarrow$ \\C-STANCE 
$\rightarrow$ Py150 $\rightarrow$ MeetingBank $\rightarrow$ ScienceQA }\\
        \midrule
         3&  SuperNI benchmark& \makecell{task1572 $\rightarrow$ task363 $\rightarrow$ task1290 $\rightarrow$ task181 $\rightarrow$ task002 $\rightarrow$ task1510 $\rightarrow$ \\ task073 $\rightarrow$ task748 $\rightarrow$ task511 $\rightarrow$ task591 $\rightarrow$ task195 $\rightarrow$ task875}\\
         \midrule
         4&  SuperNI benchmark& \makecell{task748 $\rightarrow$ task073 $\rightarrow$ task1572 $\rightarrow$ task195 $\rightarrow$ task591 $\rightarrow$ task363 $\rightarrow$ \\task1510 $\rightarrow$ task181 $\rightarrow$ task511 $\rightarrow$ task002 $\rightarrow$ task1290 $\rightarrow$ task875}\\
         \bottomrule
    \end{tabular}
    \caption{Task sequence of different task orders.}
    \label{tab:task_order}
\end{table*}

\end{document}